\pdfoutput=1
\documentclass[conference,A4]{IEEEtran}
\IEEEoverridecommandlockouts
\usepackage{cite}
\usepackage{amsmath,amssymb,amsfonts}
\usepackage{algorithmic}
\usepackage{graphicx}
\usepackage{textcomp}
\usepackage{xcolor}
\usepackage{multirow}
\usepackage{caption}
\usepackage[skins]{tcolorbox} 
\usepackage{multicol}
\usepackage{siunitx, makecell}
\usepackage{mathtools}
\usepackage{tikz}
\usetikzlibrary{arrows.meta,arrows, positioning}
\usepackage{graphics} 
\usepackage{epsfig} 
\usepackage{mathptmx} 
\usepackage{times} 
\usepackage{amsmath} 
\usepackage{amssymb}  
\usepackage{graphicx}
\usepackage{epstopdf}
\usepackage{pdfpages}
\usepackage{cite}
\usepackage{subfigure}
\usepackage[hidelinks]{hyperref}
\usepackage{amsbsy}
\usepackage{bm}
\usepackage{dblfloatfix}
\usepackage{lipsum}
\usepackage{url}
\usepackage{accents}

\makeatletter
\let\old@ps@headings\ps@headings
\let\old@ps@IEEEtitlepagestyle\ps@IEEEtitlepagestyle
\def\confheader#1{%
	\def\ps@IEEEtitlepagestyle{
		\old@ps@IEEEtitlepagestyle
		\def\@oddhead{\strut\hfill#1\hfill\strut}
		\def\@evenhead{\strut\hfill#1\hfill\strut}
	}
	\ps@headings
}
\makeatother
\confheader{
	\small{Proceedings of the 12th RSI International Conference on Robotics and Mechatronics (ICRoM 2024), Dec. 17-19, 2024, Tehran, Iran} 
}
\usepackage{eso-pic}
\newcommand{\placetextbox}[3]{
	\setbox0=\hbox{#3}
	\AddToShipoutPictureFG*{ \put(\LenToUnit{#1\paperwidth},\LenToUnit{#2\paperheight}){\vtop{{\null}\makebox[0pt][c]{#3}}}
	}
}
\placetextbox{.23}{0.055}{\textbf{\small{979-8-3315-2973-4/24/\$31.00 ©2024 IEEE}}}

\usepackage{hyperref}
\hypersetup{
	colorlinks=true,
	linkcolor=black,
	citecolor=black,
	filecolor=black,
	urlcolor=black,
}
\def\BibTeX{{\rm B\kern-.05em{\sc i\kern-.025em b}\kern-.08em
		T\kern-.1667em\lower.7ex\hbox{E}\kern-.125emX}}

\makeatletter 
\newcommand{\linebreakand}{%
  \end{@IEEEauthorhalign}
  \hfill\mbox{}\par
  \mbox{}\hfill\begin{@IEEEauthorhalign}
}
\makeatother 

\begin{document}

\title{
Control of a Biomimetic Robotic Finger Using Antagonistic Twisted and Coiled Polymer Actuators: Simulation and Experimental Validation

}

\author{\IEEEauthorblockN{Mahdi Nozari}
\IEEEauthorblockA{\textit{Center of Advanced Systems} \\ \textit{and Technologies (CAST)} \\
\textit{School of Mechanical Engineering} \\
\textit{University of Tehran}\\
nozarim@ut.ac.ir}
\and
\IEEEauthorblockN{Aghil Yousefi-Koma}
\IEEEauthorblockA{\textit{Center of Advanced Systems} \\ \textit{and Technologies (CAST)} \\
\textit{School of Mechanical Engineering} \\
\textit{University of Tehran}\\
aykoma@ut.ac.ir}
\and
\IEEEauthorblockN{Alireza Naeini}
\IEEEauthorblockA{\textit{Center of Advanced Systems} \\ \textit{and Technologies (CAST)} \\
\textit{School of Mechanical Engineering} \\
\textit{University of Tehran}\\
alirezaa.naaeini@ut.ac.ir}
\and
}

\maketitle

\begin{abstract}
The advancement of smart materials has contributed to their expanded role in robotics, despite their typically low operational frequency. Examples of these materials include SMA wires and TCPs, which are activated by heat and require cooling to return to their original state without external forces. This study introduces a new mechanism for the Surena-V humanoid robot’s finger, inspired by the function of the human hand. The design incorporates two TCP actuators and tendons to control the flexion and extension motion of the finger joint in contrast to studies where a spring returned mechanism is cooled with air flow or water to allow the actuator to expand. The force to voltage transfer function of the TCP actuators were established. Additionally, a PD controller was implemented to regulate the joint angle, with real-time angle measurements recorded via a potentiometer. The resulting data demonstrated that utilizing two antagonistic actuators can effectively control the joint angle without steady-state error or oscillations, without the need to provide cooling or spring return.
\end{abstract}


\section{Introduction}

Robotics was originally developed with the aim of producing human performance through machines. Grippers and artificial hands are the most essential parts of a robot because they are the mediators between the robot and the environment. During the development of robotics, it is easy to see that one of the most essential tasks that robots are supposed to do is grasping. This is due to the inherent human tendency to transfer some human capabilities to robots. There have been many ways to construct a hand or upper limb to create a proper grip that allows robots to grasp objects correctly and easily.

Many robotics experts have started exploring systems that could serve as suitable substitutes for human tissue. Drawing inspiration from biological systems and the advancement of smart materials, a new robotics field known as "soft robotics," which relies on soft materials and deformable structures, is progressing towards practical applications \cite{pfeil2020worm}. However, unlike natural muscles, these soft materials struggle to produce large movements under stress while maintaining durability \cite{yasa2023overview}. The hardness of materials used in soft robots typically falls between $10^4$ and $10^9$ pascals, similar to the skin or muscle tissue\cite{rus2015design}.

In contrast to rigid robots with limited degrees of freedom, soft robots can continuously deform due to their flexible structures, giving them a much higher degree of freedom \cite{bahramzadeh2014review}. Three primary methods power soft robots: fluid actuation, electrically actuated polymers, or variable-length tendons.

Fluid actuators adjust the fluid pressure inside the robot's channels, causing it to alter shape \cite{bahramzadeh2014review}. Twisted and Coiled Polymer (TCP) muscles, such as those made from nylon, are linear actuators that respond to temperature changes. Their ability to expand or contract in a limited temperature range is due to their high negative coefficient of thermal expansion (CTE) and their helical design \cite{karami2017modeling}. Polymer actuators, which are compatible with biological tissues, change size or shape when exposed to an electric field \cite{kim2007electroactive}. Variable-length tendons, available as tension cables or Shape Memory Alloy (SMA) actuators, are often embedded in flexible parts of the robot. These actuators apply controlled forces to shape different sections as needed \cite{calisti2011octopus}.

The discovery and development of twisted and coiled polymer (TCP) actuators by Haines et al. \cite{haines2014artificial} have garnered significant attention from researchers. Studies revealed that these actuators, made from low-cost materials like fishing line or sewing thread, can achieve large strain (over 30\%), high stress (over 32 MPa), and long durability (up to 1 million cycles) \cite{haines2016new}. TCP actuators also deliver high mechanical power output, with a specific work of 5.3 kW/kg, which is 17 times greater than that of mammalian skeletal muscle, which generates 0.322 kW/kg \cite{haines2014artificial}. Other benefits include affordability, high linearity, ease of fabrication, and simple actuation \cite{tang2019general}.

These benefits have made these actuators an ideal choice for a range of applications in fields such as medicine and robotics. They have already been implemented in robotic claws \cite{arefi2023design}, robotic skin \cite{matharu2024single} and robotic fingers.

Robotic claws and fingers have been developed using nylon-based TCP actuators. Wu et al. \cite{wu2017compact} designed a low-cost robotic claw driven by a nylon smart actuator, utilizing 3D printing to create a lightweight structure.

Similarly, Chu et al. \cite{yip2015high} worked on a robotic finger powered by a TCP actuator, also using 3D printing. They explored various coiling configurations for multiple actuators, achieving a maximum length change of 17.07\%.

Although many robotic hands have been built using TCP actuators, however, controlling these actuators precisely and effectively remains a challenge. 

Jafarzadeh et al.\cite{jafarzadeh2018control} implemented a Takagi-Sugeno-Kang fuzzy controller to manage the force of a nylon actuator. Using discrete-time modeling, they predicted errors to estimate the state-space parameters and determine the model’s order. A traditional proportional-integral (PI) controller was initially applied to regulate muscle force, followed by the fuzzy controller to enhance operating speed. Their findings highlighted the advantages of the fuzzy controller over the PI controller in performance.

TCP actuators expand or contract when heated due to their positive axial and negative radial thermal expansion coefficients \cite{haines2016new}, with this effect amplified by their helical structure. The strain and force produced by these actuators have a linear relationship with the applied temperature \cite{yip2017control}, making temperature a feedback signal for closed-loop control. However, directly measuring the temperature is challenging because TCPs, especially those with small diameters, have low thermal mass \cite{haines2018closed}. To address this, some researchers designed closed-loop control systems using position feedback to bypass the need for temperature measurement \cite{suzuki2018displacement}. Others monitored the temperature through the resistance change of a metal wire twisted with the fibers during actuation \cite{tang2019temperature}. This approach eliminated the need for additional sensors and provided an integrated system for both actuation and temperature feedback. However, friction between the actuator and heating wire can cause non-linearity between temperature and strain, and uneven heating may occur as the temperature is higher near the metal wire \cite{tahara2019rotational}.

Singh et al.\cite{singh2023exploring} demonstrate that the combined use of active cooling and PID control enhances the actuator’s speed and accuracy, making it more viable for practical robotic applications where quick response and reliability are essential

The need for human-robot interaction in advanced robotics motivates the use of soft and compliant systems such as TCP actuators. In addition, the actuators offer a great force-to-weight ratio, which allows for more complex hand designs with extended degrees of freedom. This research uses a bio-mimetic antagonistic actuator mechanism similar to muscles in the human body that uses soft actuators to control the movement of the finger of the Surena-V humanoid robot. Unlike earlier methods that relied on spring returns or tendon cooling, this design employs two actuators per joint to control finger movements in both directions with a single tendon. This approach enhances the performance and control accuracy of joint angles in comparison to using a single actuator. Moreover, substituting soft actuators for electric motors in the Surena-V hand not only reduces its overall weight but will also help the robot interact with it's environment more safely because of the inherent compliance of the mechanism.
\renewcommand{\arraystretch}{1.3}
\setlength{\arrayrulewidth}{0.2mm}
\setlength{\tabcolsep}{18pt}
\section{Materials \& Methods}

\subsection{Actuator}
\begin{figure}[b]
     \centering
     \includegraphics[clip,width=\linewidth]{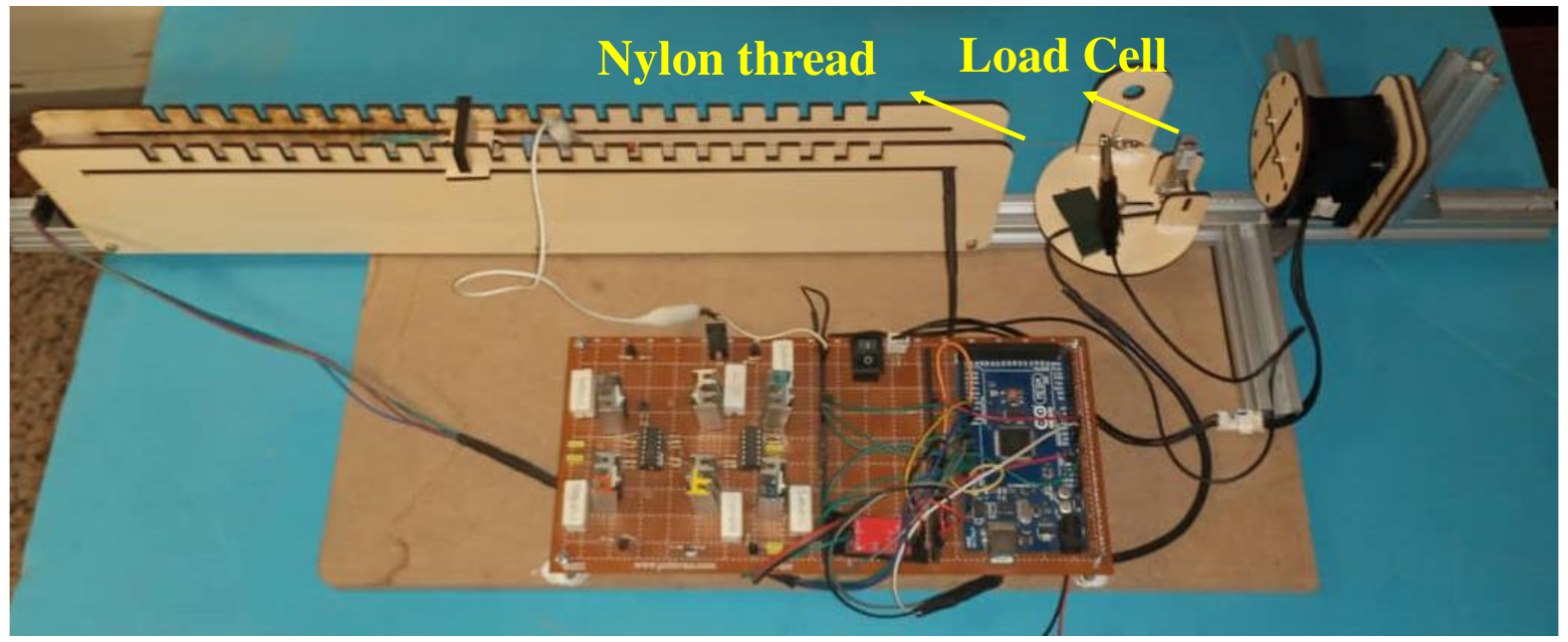}
     \caption{System identification setup used for TCP actuators}
     \label{fig:identification}
\end{figure}

TCPs initially take the form of yarn, so the first step in converting them into actuators involves shaping them into springs. This phase is broken down into two sub-steps: twisting and coiling, collectively referred to as the coil process.

In the twisting phase, the thread is twisted until it almost begins forming loops. As twisting continues, the thread then transforms into stacked loops. At this point, the opposite end of the thread must be held in place to allow the coiling process to occur.

\begin{figure*}[htbp]
		\centering
		\resizebox{\linewidth}{!}{
		\begin{tikzpicture}[auto, node distance=1cm]

		\tikzstyle{block} = [draw=black, thick, fill=blue!10, rectangle,  minimum height=3em, minimum width=6em]
		\tikzstyle{largeblock} = [draw=black ,thick , fill=blue!10, rectangle, minimum height=10em, minimum width=14em] 
		\tikzstyle{sum} = [draw, fill=black!20, circle, node distance=2cm]
		\tikzstyle{input} = [coordinate]
		\tikzstyle{output} = [coordinate]
		\tikzstyle{pinstyle} = [pin edge={to-,thin,black}]
		
		\node [input, name=input] {};
		\node [sum, right of=input] (sum) {};
		\node [block, right of=sum, node distance=2.5cm] (sel) {Actuator Selection};
		\node [block, right of=sel, node distance=4cm] (pid1) {PD Controller};
		\node [block, below of=pid1, node distance=2cm] (pid2) {PD Controller};
		\node [block, right of=pid1, node distance=3cm] (act1) {Actuator 1};
		\node [block, right of=pid2, node distance=3cm] (act2) {Actuator 2};
		
		\node [largeblock, right of=act1, node distance=5cm, yshift=-1cm, inner sep=0pt] (finger) {
			\begin{minipage}{12em} 
				\centering
				\includegraphics[width=12em]{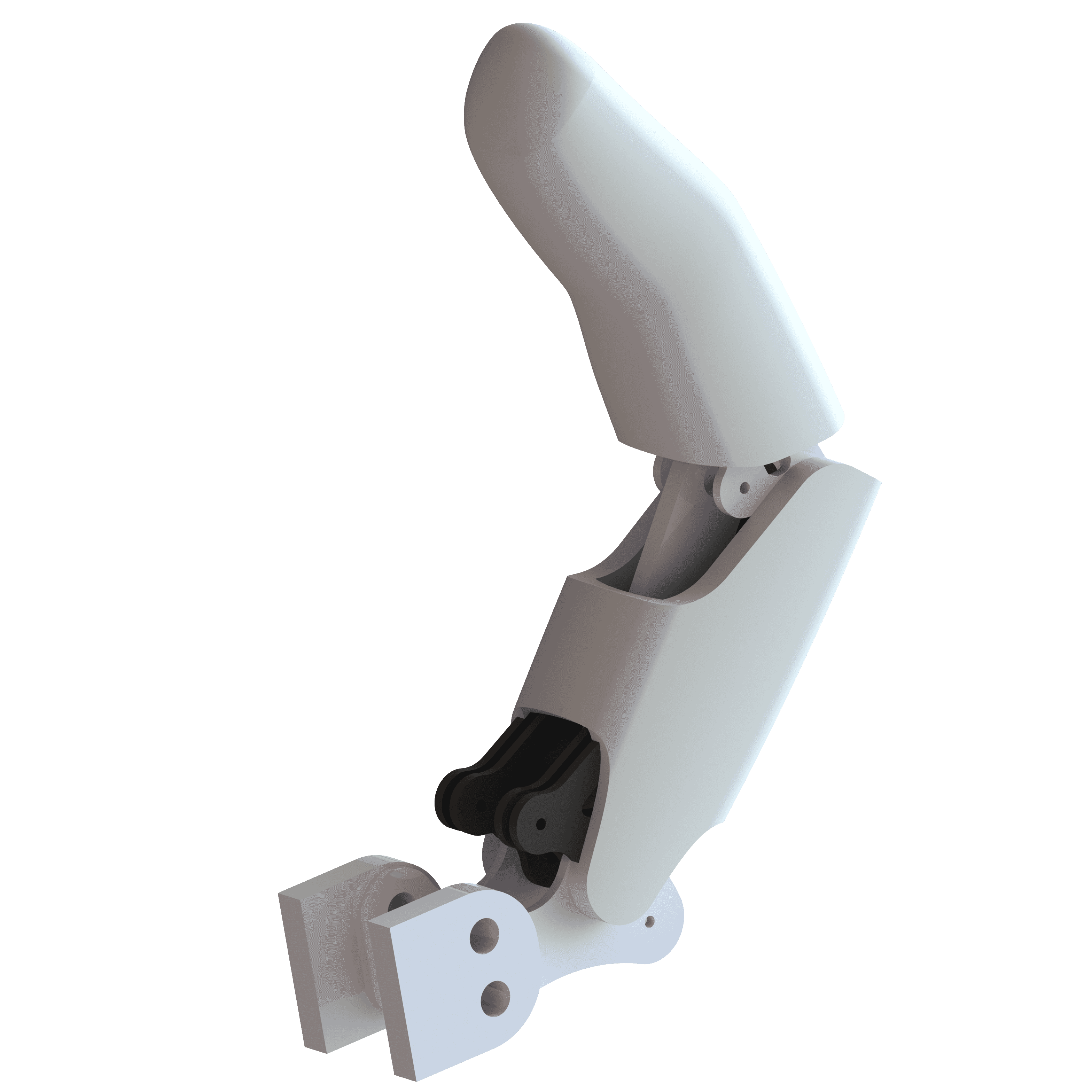} \\ 
				\vspace{0.2em}
				Finger Mechanism
			\end{minipage}
		};
		
		\draw [->] (input) -- node[pos=0.2] {Reference Angle} (sum);
		\draw [->] (sum) -- node {} (sel);
		\draw [->] (sel.east) -- ++(0.5,0) coordinate (mid) |- (pid1.west);
		\draw [->] (mid) |- (pid2.west);
		\draw [->] (pid1) -- node {$V_1$} (act1);
		\draw [->] (act1) -- node {$F_1$} (finger);
		\draw [->] (pid2) -- node {$V_2$} (act2);
		\draw [->] (act2) -- node {$F_2$} (finger);
		\draw [->] (finger.south) -- ++ (0,-1) -| node[pos=0.25] {Angle Feedback} (sum);
		
		\node at ([yshift=0.3cm]sum.west) {+};
		\node at ([yshift=-0.3cm, xshift=0.3cm]sum.south) {--}; 
		
		\end{tikzpicture}
		}
	\caption{Block diagram of the system}
        \label{fig:block_diagram}
\end{figure*}

To stabilize the initial shape of nylon threads, a thermal process is required. One method for this process involves applying voltage to both ends of the nylon thread, specifically for threads with a conductive coating. When voltage is applied, electric current flows through the thread, producing heat due to its resistance. This method provides the advantages of control and versatility, enabling the application of constant or variable heat as needed\cite{abbaszadeh}.

To model this system, the relationship between its input and output was analyzed. This method eliminates the need for a physical model and is useful for phenomena with complex physics.
\autoref{fig:identification} illustrates the identification setup. Based on the circuit design, the input parameter is the PWM value applied by the microcontroller board. This PWM signal results in a potential difference across the actuator, causing an electric current that generates heat. The temperature rise produces force and displacement in the actuator, which the load cell records. Thus, the input can be seen as the voltage produced by the PWM application, while the  output is the force generated by the operator. Consequently, the transfer function that converts power to voltage is fitted to the data as equation \autoref{eq:transfer_function}, with an accuracy of 88.02\%.

\begin{equation}
\label{eq:transfer_function}
   \frac{F(s)}{V(s)} = \frac{0.05701}{s + 0.9424}
\end{equation}

\subsection{Mechanical System}
\begin{figure}[t]
 \centering
 \includegraphics[width=\linewidth,]{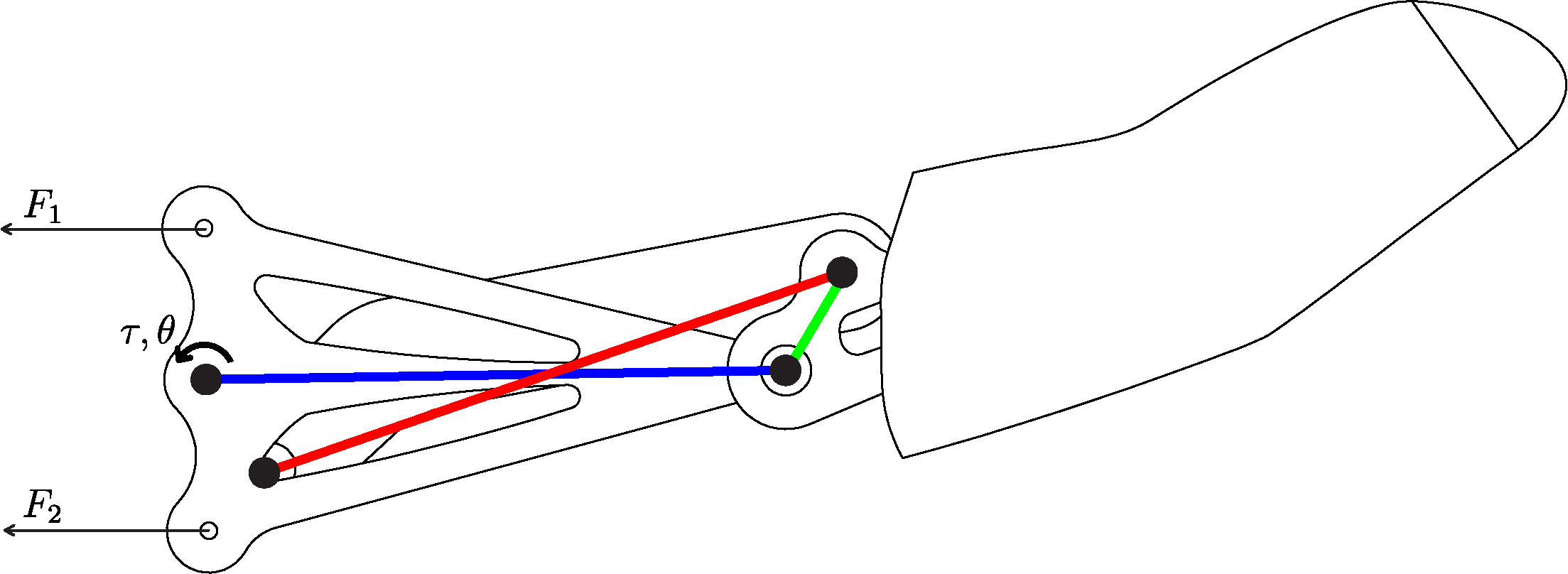}
 \caption{2D drawing of the finger mechanism. The four-bar mechanism shown with colored lines}
 \label{fig:finger_drawing}
\end{figure}

The Surena-V finger design is used as the mechanical platform in this study. In the hand of this humanoid robot, each finger has a single DoF except the thumb, which has two. The single DoF fingers have two joints in a four-bar mechanism, which can be seen in detail in \autoref{fig:finger_drawing}. The actuated joint is connected to two TCPs, which pull the link back and forth. The rotation of the link then leads to finger extension or flexion.
The system is designed in a way that the forces create the maximum torque in the finger's range of motion, that is, the effective angle of the forces is $90^\circ$. The final torque acting on the first joint in the absence of an external force and friction can be written as:
\begin{equation}
    \tau = F_1 d - F_2 d
\end{equation}
Where $F_i$ is the force created by the actuators, and $d$ is the arm length. It is worth noting that the angle feedback comes from this joint, so there isn't need for any further formulation for the position control of this angle.
\subsection{Electronics}
The TCP actuator is activated through joule heating as they are made from conductive strings. This simplifies the actuation process as there are no other heating sources needed. However, without any limitations, the consumed power in the actuator will increase the temperature without any restrictions, causing it to deviate from the identified behaviour and even to the point of failure. Moreover, heating the actuator more than needed leads to a short life cycle as the silver coating on the nylon threads is not designed to withstand high currents and will burn. To prevent this and prolong the actuator's life, a current control circuit is used to keep the temperature from rising beyond the $40 ^\circ C$ limit obtained by experience.
\begin{figure}[htbp]
     \centering
     \includegraphics[clip,width=\linewidth]{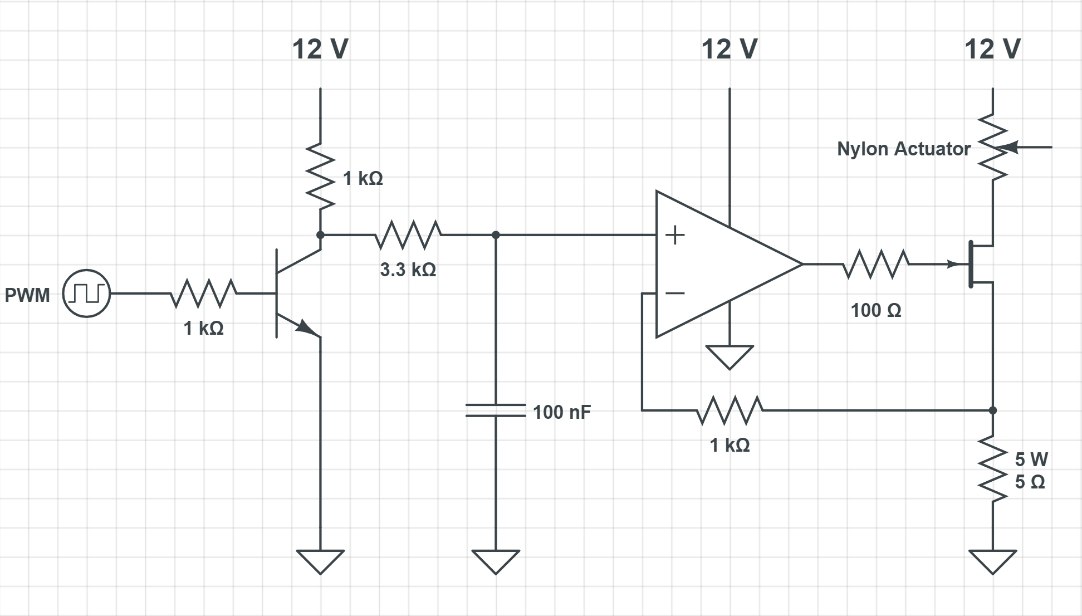}
     \caption{Current control circuit}
     \label{fig:current_control}
\end{figure}

\begin{figure*}[t]
		\centering
		\resizebox{\textwidth}{!}{
			\begin{tikzpicture}[auto, node distance=2cm, -latex']
				
				\tikzstyle{block} = [draw, thick, fill=red!20, rectangle, minimum height=4em, minimum width=6em]
				\tikzstyle{largeblock} = [draw, thick, fill=red!20, rectangle, minimum height=10em, minimum width=14em]
				\tikzstyle{pinstyle} = [pin edge={to-,thin,black}]
				
				\node [block] (computer) {Computer};
				\node [block, right=of computer] (microcontroller) {
					\begin{minipage}{8em} 
						\centering
						\includegraphics[width=8em]{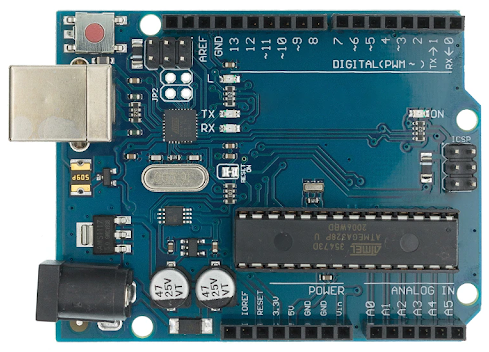} \\ 
						\vspace{0.2em} 
						Microcontroller
					\end{minipage}
				};;
				\node [block, above of=microcontroller, node distance = 3cm] (powersupply) {Power Supply};
				\node [block, right=of microcontroller] (currentcontrol1) {Current Control Circuit};
				\node [block, right=of currentcontrol1] (actuator1) {Actuator 1};
				\node [largeblock, right of=act1, node distance=8cm, yshift=-1.5cm, inner sep=0pt] (finger) {
					\begin{minipage}{12em} 
						\centering
						\includegraphics[width=12em]{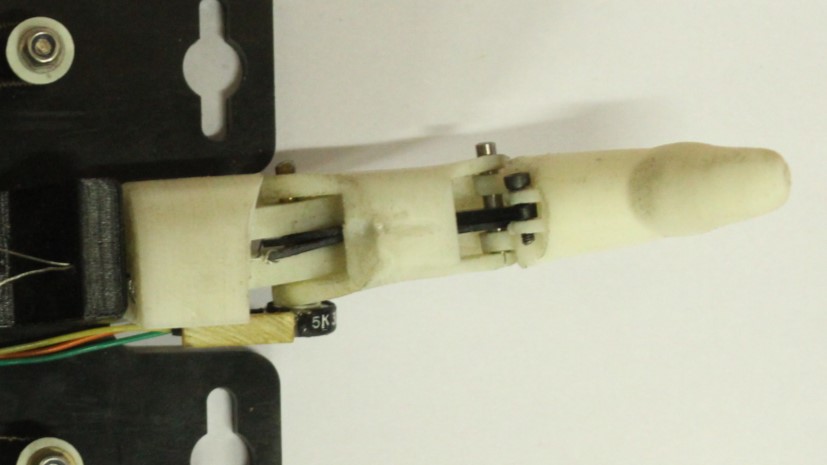} \\ 
						\vspace{0.2em}
						Finger
					\end{minipage}
				};
				\node [block, below of=finger, node distance=4cm] (potentiometer){
					\begin{minipage}{6em} 
						\centering
						\includegraphics[width=6em]{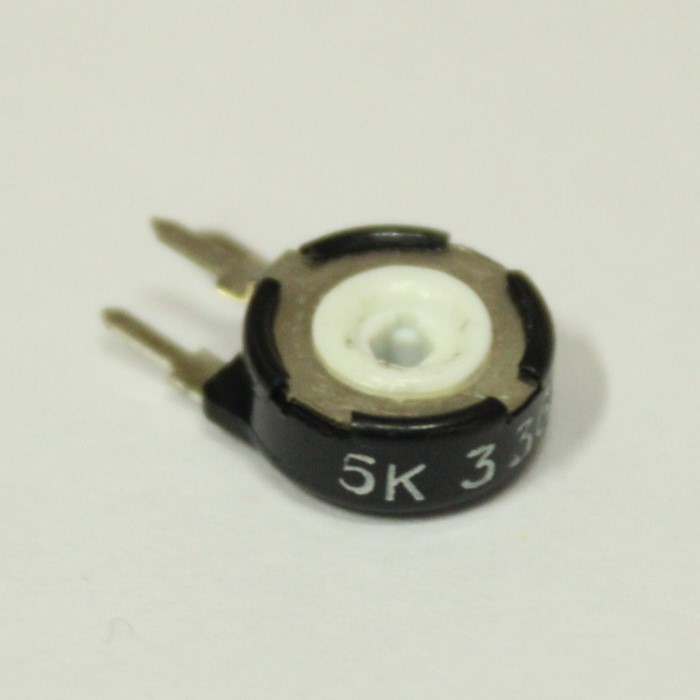} \\ 
						\vspace{0.2em}
						Potentiometer
					\end{minipage}
				};
				\node [block, below=of currentcontrol1] (currentcontrol2) {Current Control Circuit};
				\node [block, right=of currentcontrol2] (actuator2) {Actuator 2};
				\node [block, below of=microcontroller, node distance=5.5cm ] (adc) {10-bit ADC};
				
				
				\draw[->] (powersupply.east) -- ++(2,0) |- ([yshift=1cm]currentcontrol1.north) -- (currentcontrol1.north) node[midway, above right] {0-12 V};
				\draw[->] (powersupply.east) -- ++(2,0) |- ([yshift=1cm]currentcontrol2.north) -- (currentcontrol2.north) node[midway, above left] {0-12 V};
				
				\draw[->] (computer) -- node {Serial Com.} (microcontroller);
				\draw[->] (microcontroller.east) -- ++(0.5,0) |- ([xshift=-1cm]currentcontrol1.west) -- (currentcontrol1.west) node[midway, above left] {0-5 V};
				\draw[->] (microcontroller.east) |- (7,-0.5) |- ([xshift=-1cm]currentcontrol2.west) -- (currentcontrol2.west) node[below left] {0-5 V};
				\draw[->] (currentcontrol1) -- (actuator1);
				\draw[->] (actuator1) -- (finger);
				\draw[->] (finger) -- (potentiometer);
				\draw[->] (potentiometer) -- node {0-5 V signal} (adc);
				\draw[->] (adc) -- node {0-1023 } (microcontroller);
				\draw[->] (currentcontrol2) -- (actuator2);
				\draw[->] (actuator2) -- (finger);
				
			\end{tikzpicture}
			
		}
		\caption{System overview of the experimental setup}
            \label{fig:system_diagram}
	\end{figure*}

\subsection{Control}
   Although more advanced controllers show better performance than conventional controllers, the latter are able to fulfill the need for this particular study as the control method is not the main concern here. Moreover, the performance of such a controller will be tested in simulation before moving on to experiment. Using a PD controller, the system can be controlled effectively as there won't be any steady-state errors because two actuators are present. If there is to be an error in the angle, the opposite side will have it's actuator activated, compensating for the existing error. \\
    Similar to human muscles, TCP artificial muscles can generate force in only one direction, through contraction. This is also the reason behind the concept of antagonistic muscle pairs. Because of  this, after the error signal is calculated, the proper actuator needs to be selected for actuation based on the direction of the angle error. This error is then given to the PD controller, which calculates the control input as voltage. This signal generates a force that is specified by the actuator's transfer function given in \autoref{eq:transfer_function}. This force is the input to the finger's dynamics, controlling it's motion.\\
    As mentioned, another advantage of having an antagonistic muscle pair is the ability to control joint stiffness through actuator forces. By adding an equal bias input signal to the actuators, the movement of the joint is unaffected as the extra forces create equal torque in all angles and cancel out each other. When an external force is applied to the joint and tries to displace it, one of the muscles will oppose the external force depending on it's direction while the other one eases up and vice versa if the direction of the force changes. Changing the bias force applied by the actuators results in a variable joint stiffness, which helps the robotic hand interact with the environment more effectively. However, exploring the concept further and controlling the stiffness is out of the scope of this study.

\section{Results and Discussion}

\subsection{Simulation Results}
\begin{figure}[htbp]
 \centering
 \includegraphics[clip,width=\linewidth]{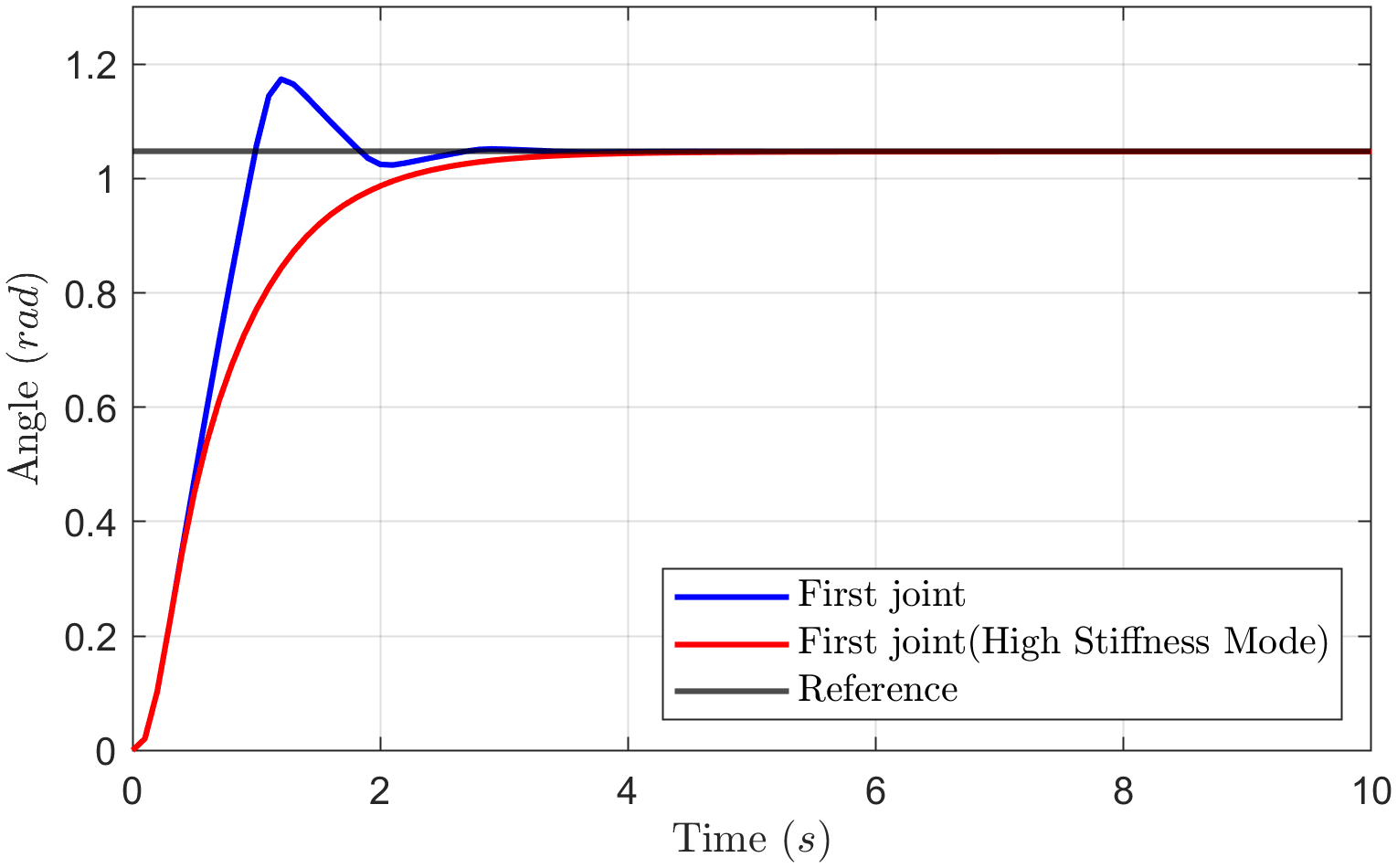}
 \caption{Simulation results for the joint angle with antagonistic actuators}
 \label{fig:simulation_results}
\end{figure}

According to the control scheme explained in \autoref{fig:block_diagram}, the control loop was simulated inside Simulink, and the mechanical system was made through Simscape using the physical properties of the real finger. Using a tuned PD controller expressed as \autoref{eq:controller}, the control input was calculated and then saturated to keep the input in the range of $0-12 V$. The acting torque can be calculated using the angle of the joint, geometrical equations, and the forces generated by the actuators. This torque is inputted to the simulated mechanical system. The system was controlled effectively, as it can be seen in \autoref{fig:simulation_results}. In addition, the control input can be manipulated to add a flat amount to both actuators, which increases the stiffness of the joint. In this mode, no overshoot can be seen, and the system will be more robust to external disturbance. 
 
    \begin{equation}
             \label{eq:controller}
        u = K_p e + K_d \dot{e}
    \end{equation}
    \begin{equation}
         K_p = 2, K_d = 1.5
    \end{equation}

\subsection{Experimental Results}

\begin{figure}[htbp]
 \centering
 \includegraphics[clip,width=\linewidth]{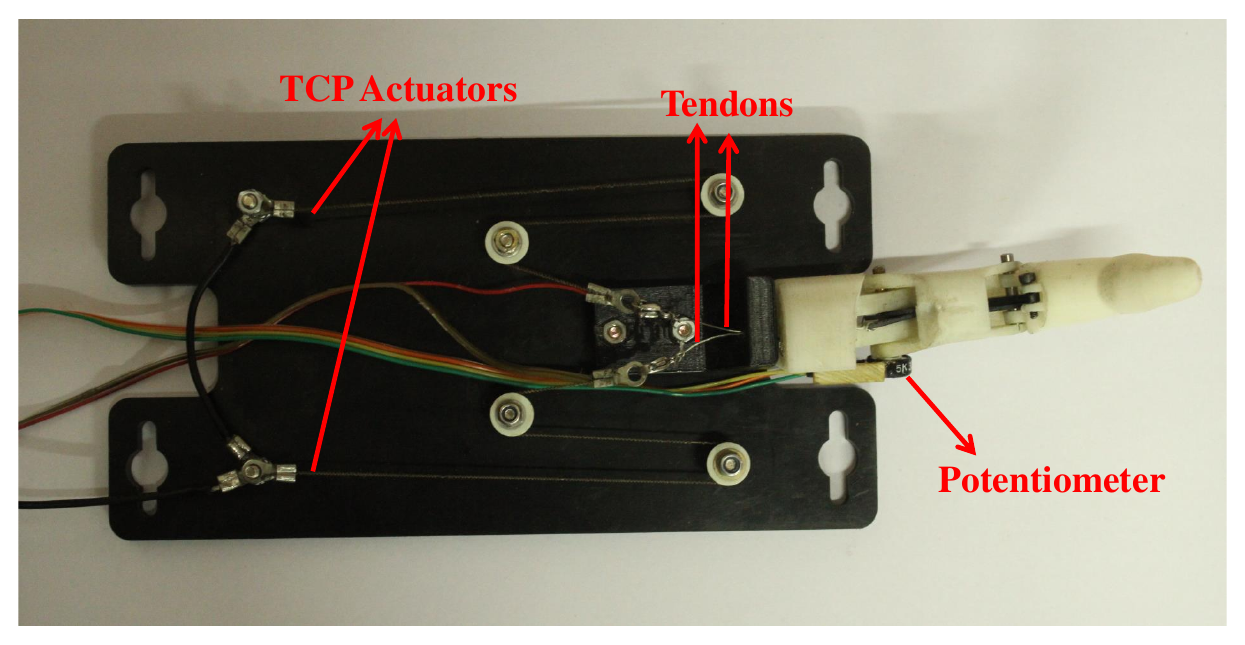}
 \caption{Mechanical setup of the antagonistic pair of TCPs connected to the index finger of Surena-V robotic hand}
 \label{fig:mechanical_setup}
\end{figure}

To validate the effectiveness of the antagonistic mechanism in experiment, a setup with one finger was built. The index finger of the Surena-V humanoid robotic hand was used with some modifications, which has a four-bar mechanism with 1 DoF as explained previously. Two tendons are connected to the link, which is directly connected to the first phalanx. The tendons are then attached to the actuators, which are guided through pulleys to keep the overall setup more compact. The TCP actuators are placed in the system in tension to keep everything intact even when the finger is open. The finger is attached to a piece that both holds the finger and leads the tendons so their effective angle is near $90^\circ$ so the maximum torque is applied with the force created. A miniature potentiometer on the joint captures the angle, which is used as feedback in the control loop. The mechanical setup can be seen in \autoref{fig:mechanical_setup}.\\

\begin{figure}[t]
 \centering
 \includegraphics[clip,width=0.45\textwidth, ]{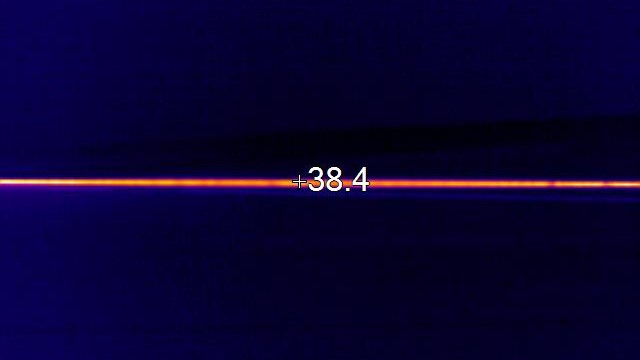}
 \caption{Steady state temperature of the TCP actuator after current control captured by a thermography camera (value in $^\circ C$)}
 \label{fig:temperature}
\end{figure}

To control the mechanism, an Arduino Uno microcontroller was used to implement a digital controller according to the diagram in \autoref{fig:system_diagram}. A current control circuit has been implemented to limit the current within the actuator, which eventually leads to a limit in the temperature of the TCP, preventing it from failing. The final temperature of a standalone actuator has been captured using a thermography camera (\autoref{fig:temperature}) to ensure that it has been limited. The PD controller specified in \autoref{eq:controller} was discretized and implemented using a sample time of $T_s = 0.1 s$ which is enough for the application as the response time of the actuators are high due to their activation by heat.\\
The experimental results confirm what has been achieved in simulation, despite some deviation due to the physical system being slower. They show that controlling the system is successful using a PD controller. However, because the actuators' behavior changes after use, the current transfer function starts to differ from \autoref{eq:transfer_function}. This results in a less responsive system, which can be seen. Interestingly, because of the nature of the mechanism, no steady-state error is recorded even with no error integration. It can be seen that while the first actuator is trying to pull the joint to the desired point, the second actuator prevents the system from high overshoot and returns the joint to the desired angle. This decreases the oscillations in the system in contrast to using a single actuator with spring-return, which will hover around the desired point.\\

\begin{figure}[htbp]
     \centering
     \includegraphics[clip,width=\linewidth]{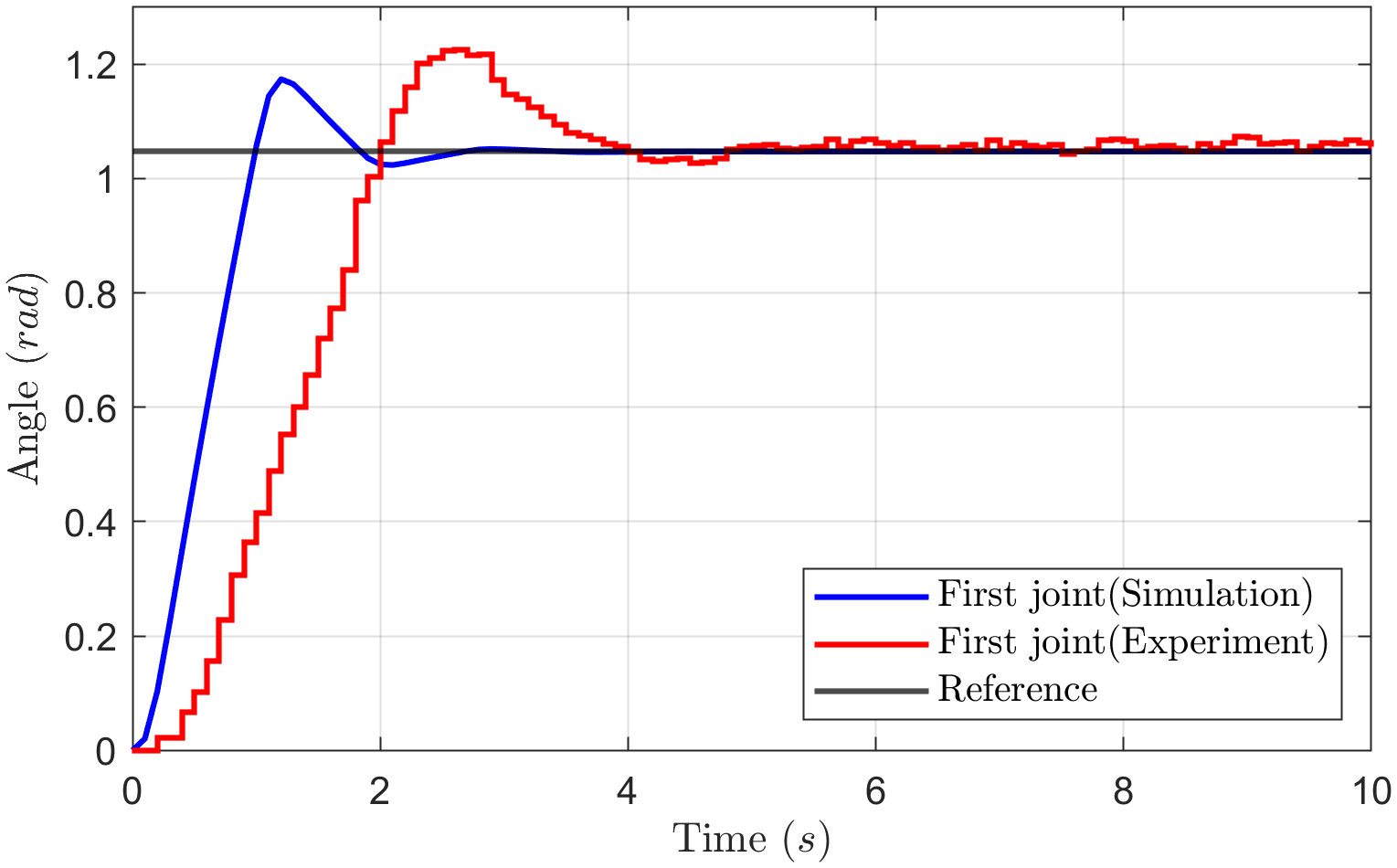}
     \caption{Simulation results for the joint angle with antagonistic actuators}
     \label{fig:experimental_results}
\end{figure}
\section{Conclusion}
This research introduced a new mechanism for controlling the finger joint of the Surena-V humanoid robot, using two TCPs as actuators. By employing a test setup, we successfully derived a transfer function that defines the relationship between the applied voltage and the resulting force exerted by the tendons. This transfer function facilitated the design of a controller aimed at enhancing joint angle control in comparison to a single actuator system.

The experimental results verified the simulation that using two actuators for bidirectional control is effective with a simple PD controller. The antagonistic configuration not only proves advantageous over single-tendon systems, as it allowed for more control over the joint angle in contrast to using a spring return system, but also facilitates joint stiffness control. The output graphs verified that the controller achieved efficient control, making the TCPs a viable choice for robotic actuation where flexibility and lightweight materials are essential.

\bibliography{bibliography}
\bibliographystyle{IEEEtran}

\end{document}